\documentclass[11pt,a4paper]{article}
\usepackage[hyphens]{url}
\usepackage[hyperref]{conll2018}
\usepackage{times}
\usepackage{latexsym}

\usepackage{amsmath}
\usepackage{algorithm}
\usepackage[noend]{algpseudocode}
\usepackage{amsmath,amsthm}
\usepackage[title]{appendix}
\usepackage{tabularx}
\usepackage{graphicx}
\graphicspath{ {images/} }
\usepackage{amssymb}
\usepackage{color}
\usepackage{placeins}

\aclfinalcopy 

\title{Evolutionary Data Measures: Understanding the Difficulty of Text Classification Tasks}

\author{Edward Collins \\
  Wluper Ltd. \\
  London, United Kingdom \\
  {\tt ed@wluper.com} \\\And
  Nikolai Rozanov \\
  Wluper Ltd. \\
  London, United Kingdom \\
  {\tt nikolai@wluper.com} \\\And
  Bingbing Zhang \\
  Wluper Ltd.\\
  London, United Kingdom \\
  {\tt bingbing@wluper.com} \\}
\date{}

\begin{document}
\maketitle
\begin{abstract}

Classification tasks are usually analysed and improved through new model architectures or hyperparameter optimisation but the underlying properties of datasets are discovered on an ad-hoc basis as errors occur. However, understanding the properties of the data is crucial in perfecting models. In this paper we analyse exactly which characteristics of a dataset best determine how difficult that dataset is for the task of text classification. We then propose an intuitive measure of difficulty for text classification datasets which is simple and fast to calculate. We show that this measure generalises to unseen data by comparing it to state-of-the-art datasets and results. This measure can be used to analyse the precise source of errors in a dataset and allows fast estimation of how difficult a dataset is to learn. We searched for this measure by training 12 classical and neural network based models on 78 real-world datasets, then use a genetic algorithm to discover the best measure of difficulty. Our difficulty-calculating code\footnote{\url{https://github.com/Wluper/edm}} and datasets\footnote{\url{http://data.wluper.com}} are publicly available.

\end{abstract}

\section{Introduction} 
\label{intro}

If a machine learning (ML) model is trained on a dataset then the same machine learning model on the same dataset but with more granular labels will frequently have lower performance scores than the original model (see results in \newcite{charCNN,socher2013recursive,generativeTextClass,fastTextSimpleClassifier,charCRNN,vdcnn}). Adding more granularity to labels makes the dataset harder to learn - it increases the dataset's difficulty. It is obvious that some datasets are more difficult for learning models than others, but is it possible to quantify this “difficulty”? In order to do so, it would be necessary to understand exactly what characteristics of a dataset are good indicators of how well models will perform on it so that these could be combined into a single measure of difficulty.

Such a difficulty measure would be useful as an analysis tool and as a performance estimator. As an analysis tool, it would highlight precisely what is causing difficulty in a dataset, reducing the time practitioners need spend analysing their data. As a performance estimator, when practitioners approach new datasets they would be able to use this measure to predict how well models are likely to perform on the dataset.


The complexity of datasets for ML has been previously examined 
\cite{geometricComplexity,geometricComplexity2,classInterferenceRef,analysisOfClassifiers}, but these works focused on analysing feature space data $\in {\rm I\!R}^n$. These methods do not easily apply to natural language, because they would require the language be embedded into feature space in some way, for example with a word embedding model which introduces a dependency on the model used. We extend previous notions of difficulty to English language text classification, an important component of natural language processing (NLP) applicable to tasks such as sentiment analysis, news categorisation and automatic summarisation \cite{socher2013recursive,newsCat,collins2017supervised}. All of our recommended calculations depend only on counting the words in a dataset. 


\subsection{Related Work}
\label{i_related}

One source of difficulty in a dataset is mislabelled items of data (noise). \newcite{identifyingMislabelled} showed that filtering noise could produce large gains in model performance, potentially yielding larger improvements than hyperparameter optimisation \cite{filterVsopt}. We ignored noise in this work because it can be reduced with proper data cleaning and is not a part of the true signal of the dataset. We identified four other areas of potential difficulty which we attempt to measure:

\paragraph{Class Interference.} Text classification tasks to predict the 1 - 5 star rating of a review are more difficult than  predicting whether a review is positive or negative \cite{charCNN,socher2013recursive,generativeTextClass,fastTextSimpleClassifier,charCRNN,vdcnn}, as reviews given four stars share many features with those given five stars. \newcite{trainingHighMulticlass} describe how as the number of classes in a dataset increases, so does the potential for "confusability" where it becomes difficult to tell classes apart, therefore making a dataset more difficult. Previous work has mostly focused on this confusability - or class interference - as a source of difficulty in machine learning tasks \cite{classInterferenceRef,firstComplexity,geometricComplexity,complexityEstimationLS,geometricComplexity2}, a common technique being to compute a minimum spanning tree on the data and count the number of edges which link different classes. 

\paragraph{Class Diversity.} Class diversity provides information about the composition of a dataset by measuring the relative abundances of different classes \cite{shannonDiversity}. Intuitively, it gives a measure of how well a model could do on a dataset without examining any data items and always predicting the most abundant class. Datasets with a single overwhelming class are easy to achieve high accuracies on by always predicting the most abundant class. A measure of diversity is one feature used by \newcite{reviewRecomBingel} to identify datasets which would benefit from multi-task learning. 

\paragraph{Class Balance.} Unbalanced classes are a known problem in machine learning \cite{classImbalIsIssue, smote}, particularly if classes are not easily separable \cite{classImbalProblem}. Underrepresented classes are more difficult to learn because models are not exposed to them as often.

\paragraph{Data Complexity.} Humans find some pieces of text more difficult to comprehend than others. How difficult a piece of text is to read can be calculated automatically using measures such as those proposed by \newcite{smog,ari,freRef}. If a piece of text is more difficult for a human to read and understand, the same may be true for an ML model.

\section{Method} 
\label{method}

We used 78 text classification datasets and trained 12 different ML algorithms on each of the datasets for a total of 936 models trained. The highest achieved macro F1 score \cite{f1Ref}, on the test set for each model was recorded. Macro F1 score is used because it is valid under imbalanced classes. We then calculated 48 different statistics which attempt to measure our four hypothesised areas of difficulty for each dataset. We then needed to discover which statistic or combination thereof correlated with model F1 scores. 


We wanted the discovered difficulty measure to be useful as an analysis tool, so we enforced a restriction that the difficulty measure should be composed only by summation, without weighting the constituent statistics. This meant that each difficulty measure could be used as an analysis tool by examining its components and comparing them to the mean across all datasets. 

Each difficulty measure was represented as a binary vector of length 48 - one bit for each statistic - each bit being 1 if that statistic was used in the difficulty measure. We therefore had $2^{48}$ possible different difficulty measures that may have correlated with model score and needed to search this space efficiently.

Genetic algorithms are biologically inspired search algorithms and are good at searching large spaces efficiently \cite{gaGoodForSearch}. They maintain a population of candidate difficulty measures and combine them based on their "fitness" - how well they correlate with model scores - so that each "parent" can pass on pieces of information about the search space \cite{immuneGA}. Using a genetic algorithm, we efficiently discovered which of the possible combinations of statistics correlated with model performance.

\subsection{Datasets}
\label{m_datasets}

\begin{table*}[!t]
\small
\begin{center}
\begin{tabularx}{\linewidth}{|p{7cm}|X|X|X|X|}
\hline \bf Dataset Name & \bf Num. Class. & \bf Train Size & \bf Valid Size & \bf Test Size \\ \hline
AG's News \cite{charCNN}                   				& 4   & 108000 & 12000 & 7600  \\
Airline Twitter Sentiment \cite{CrowdFlower}   			& 3   & 12444  & -     & 2196  \\
ATIS \cite{price1990evaluation}                  		& 26  & 9956   & -     & 893   \\
Corporate Messaging \cite{CrowdFlower}         			& 4   & 2650   & -     & 468   \\
ClassicLit                  							& 4   & 40489  & 5784  & 11569 \\
DBPedia \cite{depdia}                     				& 14  & 50400  & 5600  & 7000  \\
Deflategate \cite{CrowdFlower}                 			& 5   & 8250   & 1178  & 2358  \\
Disaster Tweets \cite{CrowdFlower}             			& 2   & 7597   & 1085  & 2172  \\
Economic News Relevance \cite{CrowdFlower}     			& 2   & 5593   & 799   & 1599  \\
Grammar and Product Reviews \cite{gprkaggle} 			& 5   & 49730  & 7105  & 14209 \\
Hate Speech \cite{davidson2017automated}         		& 3   & 17348  & 2478  & 4957  \\
Large Movie Review Corpus \cite{maas2011learning}    	& 2   & 35000  & 5000  & 10000 \\
London Restaurant Reviews (TripAdvisor\footnote{\url{https://www.kaggle.com/PromptCloudHQ/londonbased-restaurants-reviews-on-tripadvisor}})   						 	 & 5   & 12056  & 1722  & 3445  \\
New Year's Tweets \cite{CrowdFlower}           			& 10  & 3507   & 501   & 1003  \\
New Year's Tweets \cite{CrowdFlower}           			& 115 & 3507   & 501   & 1003  \\
Paper Sent. Classification \cite{pscuci}  				& 5   & 2181   & 311   & 625   \\
Political Social Media \cite{CrowdFlower}      			& 9   & 3500   & 500   & 1000  \\
Question Classification \cite{li2002learning}	        & 6   & 4906   & 546   & 500   \\
Review Sentiments \cite{kotzias2015group}               & 2   & 2100   & 300   & 600   \\
Self Driving Car Sentiment \cite{CrowdFlower}  			& 6   & 6082   & -     & 1074  \\
SMS Spam Collection \cite{smsspam}         				& 2   & 3901   & 558   & 1115  \\
SNIPS Intent Classification \cite{snipsPaper} 			& 7   & 13784  & -     & 700   \\
Stanford Sentiment Treebank \cite{socher2013recursive}	& 3   & 236076 & 1100  & 2210  \\
Stanford Sentiment Treebank \cite{socher2013recursive} 	& 2   & 117220 & 872   & 1821  \\
Text Emotion \cite{CrowdFlower}                			& 13  & 34000  & -     & 6000  \\
Yelp Reviews \cite{yelpchallenge}		                & 5   & 29250  & 3250  & 2500  \\
YouTube Spam \cite{ytsuci}								& 2   & 1363   & 194   & 391   \\
\hline
\end{tabularx}
\end{center}
\caption{\label{dataset_table} The 27 different publicly available datasets we gathered with references.}
\end{table*}

We gathered 27 real-world text classification datasets from public sources, summarised in Table \ref{dataset_table}; full descriptions are in Appendix \ref{app_a_dset_desc}.

We created 51 more datasets by taking two or more of the original 27 datasets and combining all of the data points from each into one dataset. The label for each data item was the name of the dataset which the text originally came from. We combined similar datasets in this way, for example two different datasets of tweets, so that the classes would not be trivially distinguishable - there is no dataset to classify text as either a tweet or Shakespeare for example as this would be too easy for models. The full list of combined datasets is in Appendix \ref{a_dset_comb}.

Our datasets focus on short text classification by limiting each data item to 100 words. We demonstrate that the difficulty measure we discover with this setup generalises to longer text classification in Section \ref{r_does_generalise}. All datasets were lowercase with no punctuation. For datasets with no validation set, 15\% of the training set was randomly sampled as a validation set at runtime.

\subsection{Dataset Statistics}
\label{m_metrics}

We calculated 12 distinct statistics with different n-gram sizes to produce 48 statistics of each dataset. These statistics are designed to increase in value as difficulty increases. The 12 statistics are described here and a listing of the full 48 is in Appendix \ref{a_stats} in Table \ref{statistics_table}. We used n-gram sizes from unigrams up to 5-grams and recorded the average of each statistic over all n-gram sizes. All probability distributions were count-based - the probability of a particular n-gram / class / character was the count of occurrences of that particular entity divided by the total count of all entities.

\subsubsection{Class Diversity}
\label{m_ds_class_div}

We recorded the Shannon Diversity Index and its normalised variant the Shannon Equitability \cite{shannonDiversity} using the count-based probability distribution of classes described above. 




\subsubsection{Class Balance}

We propose a simple measure of class imbalance:

\begin{align}
Imbal = \sum_{c = 1}^{C} \left| \frac{1}{C} - \frac{n_c}{T_{DATA}} \right|
\end{align}
$C$ is the total number of classes, $n_c$ is the count of items in class $c$ and $T_{DATA}$ is the total number of data points. This statistic is 0 if there are an equal number of data points in every class and the upper bound is $2 \left( 1 - \frac{1}{C} \right)$ and is achieved when one class has all the data points - a proof is given in Appendix \ref{a_stats_classimbal}.

\subsubsection{Class Interference}

Per-class probability distributions were calculated by splitting the dataset into subsets based on the class of each data point and then computing count-based probability distributions as described above for each subset.

\paragraph{Hellinger Similarity} One minus both the average and minimum Hellinger Distance \cite{hellingerDist} between each pair of classes. Hellinger Distance is 0 if two probability distributions are identical so we subtract this from 1 to give a higher score when two classes are similar giving the Hellinger Similarity. One minus the minimum Hellinger Distance is the maximum Hellinger Similarity between classes.







\paragraph{Top N-Gram Interference} Average Jaccard similarity \cite{jaccard} between the set of the top 10 most frequent n-grams from each class. N-grams entirely composed of stopwords were ignored.



\paragraph{Mutual Information} Average mutual information \cite{mutualInfoRef} score between the set of the top 10 most frequent n-grams from each class. N-grams entirely composed of stopwords were ignored.





\subsubsection{Data Complexity}

\paragraph{Distinct n-grams : Total n-grams} Count of distinct n-grams in a dataset divided by the total number of n-grams. Score of 1 indicates that each n-gram occurs once in the dataset. 

\paragraph{Inverse Flesch Reading Ease} The Flesch Reading Ease (FRE) formula grades text from 100 to 0, 100 indicating most readable and 0 indicating difficult to read \cite{freRef}. We take the reciprocal of this measure.

\paragraph{N-Gram and Character Diversity} Using the Shannon Index and Equitability described by \newcite{shannonDiversity} we calculate the diversity and equitability of n-grams and characters. Probability distributions are count-based as described at the start of this section.

\subsection{Models}
\label{m_models}

\begin{table*}[!t]
\small
\begin{center}
\begin{tabularx}{\linewidth}{|X|X|X|}
\hline 
\bf Word Embedding Based    & \bf tf-idf Based                  & \bf Character Based \\ 
\hline
LSTM-RNN                    & Adaboost                          & 3 layer CNN \\
GRU-RNN                     & Gaussian Naive Bayes (GNB)        & -           \\
Bidirectional LSTM-RNN      & 5-Nearest Neighbors               & -           \\
Bidirectional GRU-RNN       & (Multinomial) Logistic Regression & -           \\
Multilayer Perceptron (MLP) & Random Forest                     & -           \\
-                           & Support Vector Machine            & -           \\
\hline
\end{tabularx}
\end{center}
\caption{\label{models_table} Models summary organised by which input type they use.}
\end{table*}

To ensure that any discovered measures did not depend on which model was used (i.e. that they were model agnostic), we trained 12 models on every dataset. The models are summarised in Table \ref{models_table}. Hyperparameters were not optimised and were identical across all datasets. Specific implementation details of the models are described in Appendix \ref{app_b_model_desc}. Models were evaluated using the macro F1-Score. These models used three different representations of text to learn from to ensure that the discovered difficulty measure did not depend on the representation. These are:

\paragraph{Word Embeddings} Our neural network models excluding the Convolutional Neural Network (CNN) used 128-dimensional FastText \cite{fasttextRef} embeddings trained on the One Billion Word corpus \cite{billionWordCorpus} which provided an open vocabulary across the datasets.

\paragraph{Term Frequency Inverse Document Frequency (tf-idf)} Our classical machine learning models represented each data item as a tf-idf vector \cite{tfidfRef}. This vector has one entry for each word in the vocab and if a word occurs in a data item, then that position in the vector is the word's tf-idf score.

\paragraph{Characters} Our CNN, inspired by \newcite{charCNN}, sees only the characters of each data item. Each character is assigned an ID and the list of IDs is fed into the network.

\subsection{Genetic Algorithm}
\label{m_ga}

The genetic algorithm maintains a \textit{population} of candidate difficulty measures, each being a binary vector of length 48 (see start of Method section). At each time step, it will evaluate each member of the population using a \textit{fitness function}. It will then select pairs of parents based on their fitness, and perform \textit{crossover} and \textit{mutation} on each pair to produce a new child difficulty measure, which is added to the next population. This process is iterated until the fitness in the population no longer improves.

\paragraph{Population} The genetic algorithm is non-randomly initialised with the 48 statistics described in Section \ref{m_metrics} - each one is a difficulty measure composed of a single statistic. 400 pairs of parents are sampled with replacement from each population, so populations after this first time step will consist of 200 candidate measures. The probability of a measure being selected as a parent is proportional to its fitness. 

\paragraph{Fitness Function} The fitness function of each difficulty measure is based on the Pearson correlation \cite{pearsonCorrelation}. Firstly, the Pearson correlation between the difficulty measure and the model test set score is calculated for each individual model. The Harmonic mean of the correlations of each model is then taken, yielding the fitness of that difficulty measure. Harmonic mean is used because it is dominated by its lowest constituents, so if it is high then correlation must be high for every model.

\paragraph{Crossover and Mutation} To produce a new difficulty measure from two parents, the constituent statistics of each parent are randomly intermingled, allowing each parent to pass on information about the search space. This is done in the following way: for each of the 48 statistics, one of the two parents is randomly selected and if the parent uses that statistic, the child also does. This produces a child which has features of both parents. To introduce more stochasticity to the process and ensure that the algorithm does not get trapped in a local minima of fitness, the child is mutated. Mutation is performed by randomly adding or taking away each of the 48 statistics with probability 0.01. After this process, the child difficulty measure is added to the new population.

\paragraph{Training} The process of calculating fitness, selecting parents and creating child difficulty measures is iterated until there has been no improvement in fitness for 15 generations. Due to the stochasticity in the process, we run the whole evolution 50 times. We run 11 different variants of this evolution, leaving out different statistics of the dataset each time to test which are most important in finding a good difficulty measure, in total running 550 evolutions. Training time is fast, averaging 79 seconds per  evolution with a standard deviation of 25 seconds, determined over 50 runs of the algorithm on a single CPU. 

\section{Results and Discussion} 
\label{results}

\begin{figure}[!htb]
\includegraphics[width=\linewidth]{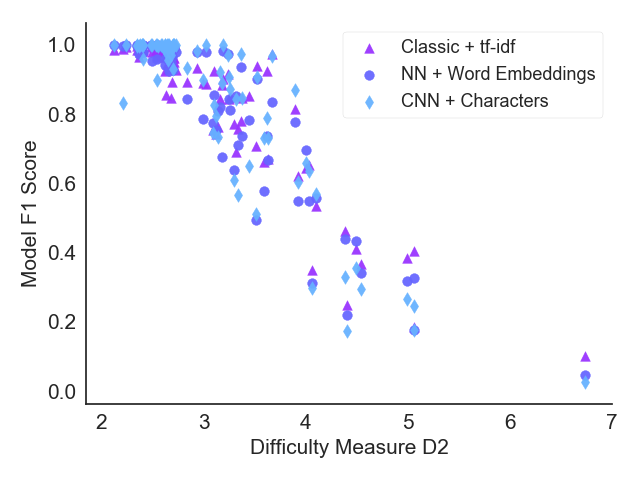}
\caption{\label{fig_model_scores} Model F1 scores against difficulty measure D2 for each of the three input types.}
\end{figure}

The four hypothesized areas of difficulty - Class Diversity, Balance and Interference and Data Complexity - combined give a model agnostic measure of difficulty. All runs of the genetic algorithm produced different combinations of statistics which had strong negative correlation with model scores on the 78 datasets. The mean correlation was $-0.8795$ and the standard deviation was $0.0046$. Of the measures found through evolution we present two of particular interest:

\begin{enumerate}

\item \textbf{D1:} \textit{Distinct Unigrams : Total Unigrams + Class Imbalance + Class Diversity + Top 5-Gram Interference + Maximum Unigram Hellinger Similarity + Unigram Mutual Info}. This measure achieves the highest correlation of all measures at $-0.8845$.

\item \textbf{D2:} \textit{Distinct Unigrams : Total Unigrams + Class Imbalance + Class Diversity + Maximum Unigram Hellinger Similarity + Unigram Mutual Info}. This measure is the shortest measure which achieves a higher correlation than the mean, at $-0.8814$. This measure is plotted against model F1 scores in Figure \ref{fig_model_scores}. 

\end{enumerate}

We perform detailed analysis on difficulty measure D2 because it relies only on the words of the dataset and requires just five statistics. This simplicity makes it interpretable and fast to calculate. All difficulty measures which achieved a correlation better than $-0.88$ are listed in Appendix \ref{app_further_evolution_results}, where Figure \ref{fig_selection_freq} also visualises how often each metric was selected.

\subsection{Does it Generalise?}
\label{r_does_generalise}

\begin{table*}[!t]
\small
\begin{center}
\begin{tabularx}{\linewidth}{p{4cm}XXXXXXXX|X}
\hline 
\bf Model & \bf AG & \bf Sogou & \bf Yelp P. & \bf Yelp F. & \bf DBP & \bf Yah A. & \bf Amz. P. & \bf Amz. F. & \bf Corr. \\ 
\hline

\textcolor{red}{\bf D2} & \textcolor{red}{\bf 3.29} & \textcolor{red}{\bf 3.77} & \textcolor{red}{\bf 3.59} & \textcolor{red}{\bf 4.42} & \textcolor{red}{\bf 3.50} & \textcolor{red}{\bf 4.51} & \textcolor{red}{\bf 3.29} & \textcolor{red}{\bf 4.32} & - \\
char-CNN \cite{charCNN} & 87.2 & 95.1 & 94.7 & 62 & 98.3 & 71.2l & 95.1 & 59.6 & -0.86 \\
Bag of Words \cite{charCNN} & 88.8 & 92.9 & 92.2 & 57.9 & 96.6 & 68.9 & 90.4 & 54.6 & -0.87 \\
Discrim. LSTM \cite{generativeTextClass} & 92.1 & 94.9 & 92.6 & 59.6 & 98.7 & 73.7 & - & - & -0.87 \\
Genertv. LSTM \cite{generativeTextClass} & 90.6 & 90.3 & 88.2 & 52.7 & 95.4 & 69.3 & - & - & -0.88 \\
Kneser-Ney Bayes \cite{generativeTextClass} & 89.3 & 94.6 & 81.8 & 41.7 & 95.4 & 69.3 & - & - & -0.79 \\
FastText Lin. Class. \cite{fastTextSimpleClassifier} & 91.5 & 93.9 & 93.8 & 60.4 & 98.1 & 72 & 91.2 & 55.8 & -0.86 \\
Char CRNN \cite{charCRNN} & 91.4 & 95.2 & 94.5 & 61.8 & 98.6 & 71.7 & 94.1 & 59.2 & -0.88 \\
VDCNN \cite{vdcnn} & 91.3 & 96.8 & 95.7 & 64.7 & 98.7 & 73.4 & 95.7 & 63 & -0.88 \\ \hline
\bf Harmonic Mean &  & & & & & & & & \bf -0.86 \\ 

\hline

\end{tabularx}
\end{center}
\caption{\label{generalise_table} Difficulty measure D2 compared to recent results from papers on large-scale text classification. The correlation column reports the correlation between difficulty measure D2 and the model scores for that row.}


\end{table*}

A difficulty measure is useful as an analysis and performance estimation tool if it is model agnostic and provides an accurate difficulty estimate on unseen datasets.

When running the evolution, the F1 scores of our character-level CNN were not observed by the genetic algorithm. If the discovered difficulty measure still correlated with the CNN's scores despite never having seen them during evolution, it is more likely to be model agnostic. The CNN has a different model architecture to the other models and has a different input type which encodes no prior knowledge (as word embeddings do) or contextual information about the dataset (as tf-idf does). D1 has a correlation of $-0.9010$ with the CNN and D2 has a correlation of $-0.8974$ which suggests that both of our presented measures do not depend on what model was used.

One of the limitations of our method was that our models never saw text that was longer than 100 words and were never trained on any very large datasets (i.e. $>$1 million data points). We also performed no hyperparameter optimisation and did not use state-of-the-art models. To test whether our measure generalises to large datasets with text longer than 100 words, we compared it to some recent state-of-the-art results in text classification using the eight datasets described by \newcite{charCNN}. These results are presented in Table \ref{generalise_table} and highlight several important findings.

\paragraph{The Difficulty Measure Generalises to Very Large Datasets and Long Data Items.} The smallest of the eight datasets described by \newcite{charCNN} has 120 000 data points and the largest has 3.6 million. As D2 still has a strong negative correlation with model score on these datasets, it seems to generalise to large datasets. Furthermore, these large datasets do not have an upper limit of data item length (the mean data item length in Yahoo Answers is 520 words), yet D2 still has strong negative correlation with model score, showing that it does not depend on data item length.

\paragraph{The Difficulty Measure is Model and Input Type Agnostic.} The state-of-the-art models presented in Table \ref{generalise_table} have undergone hyperparameter optimisation and use different input types including per-word learned embeddings \cite{generativeTextClass}, n-grams, characters and n-gram embeddings \cite{fastTextSimpleClassifier}. As D2 still has a strong negative correlation with these models' scores, we can conclude that it has accurately measured the difficulty of a dataset in a way that is useful regardless of which model is used.

\paragraph{The Difficulty Measure Lacks Precision.} The average score achieved on the Yahoo Answers dataset is $69.9\%$ and its difficulty is $4.51$. The average score achieved on Yelp Full is $56.8\%$, $13.1\%$ less than Yahoo Answers and its difficulty is $4.42$. In ML terms, a difference of 13\% is significant yet our difficulty measure assigns a higher difficulty to the easier dataset. However, Yahoo Answers, Yelp Full and Amazon Full, the only three of \newcite{charCNN}'s datasets for which the state-of-the-art is less than $90\%$, all have difficulty scores $>4$, whereas the five datasets with scores $>90\%$ all have difficulty scores between 3 and 4. This indicates that the difficulty measure in its current incarnation may be more effective at assigning a class of difficulty to datasets, rather than a regression-like value. 

\subsection{Difficulty Measure as an Analysis Tool}
\label{r_analysis_tool}

\begin{table}[h]
\small
\begin{center}
\begin{tabularx}{\linewidth}{|p{4.5cm}|X|X|}
\hline 
\bf Statistic & \bf Mean & \bf Sigma  \\ 
\hline

Distinct Words : Total Words      & 0.0666 & 0.0528 \\
Class Imbalance                   & 0.503  & 0.365 \\
Class Diversity                   & 0.905  & 0.759 \\
Max. Unigram Hellinger Similarity & 0.554  & 0.165 \\
Top Unigram Mutual Info           & 1.23   & 0.430 \\
\hline

\end{tabularx}
\end{center}
\caption{\label{means_table} Means and standard deviations of the constituent statistics of difficulty measure D2 across the 78 datasets from this paper and the eight datasets from \newcite{charCNN}.}
\end{table}

As our difficulty measure has no dependence on learned weightings or complex combinations of statistics - only addition - it can be used to analyse the sources of difficulty in a dataset directly. To demonstrate, consider the following dataset:

\paragraph{Stanford Sentiment Treebank Binary Classification (SST\_2) \cite{sstRef}} SST is a dataset of movie reviews for which the task is to classify the sentiment of each review. The current state-of-the-art accuracy is $91.8\%$ \cite{sentimentNeuron}.
\\


\begin{figure}[!htb]
\includegraphics[width=\linewidth]{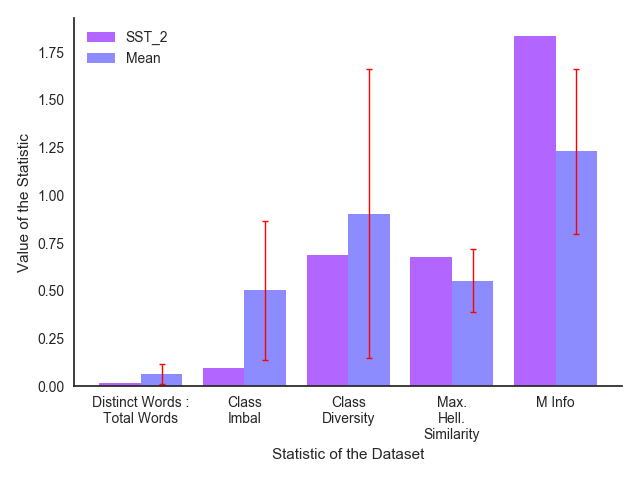}
\caption{\label{fig_analysis_tool} Constituents of difficulty measure D2 for SST, compared to the mean across all datasets.}
\end{figure}

\noindent Figure \ref{fig_analysis_tool} shows the values of the constituent statistics of difficulty measure D2 for SST and the mean values across all datasets. The mean (right bar) also includes an error bar showing the standard deviation of statistic values. The exact values of the means and standard deviations for each statistic in measure D2 are shown in Table \ref{means_table}.

Figure \ref{fig_analysis_tool} shows that for SST\_2 the Mutual Information is more than one standard deviation higher than the mean.  A high mutual information score indicates that reviews have both positive and negative features. For example, consider this review: \textit{"de niro and mcdormand give solid performances but their screen time is sabotaged by the story s inability to create interest"} which is labelled "positive". There is a positive feature referring to the actors' performances and a negative one referring to the plot. A solution to this would be to treat the classification as a multi-label problem where each item can have more than one class, although this would require that the data be relabelled by hand. An alternate solution would be to split reviews like this into two separate ones: one with the positive component and one with the negative.


Furthermore, Figure \ref{fig_analysis_tool} shows that the Max. Hellinger Similarity is higher than average for SST\_2, indicating that the two classes use similar words. Sarcastic reviews use positive words to convey a negative sentiment \cite{sarcsamSentimentAnalysis} and could contribute to this higher value, as could mislabelled items of data. Both of these things portray one class with features of the other - sarcasm by using positive words with a negative tone and noise because positive examples are labelled as negative and vice versa. This kind of difficulty can be most effectively reduced by filtering noise \cite{filterVsopt}.

To show that our analysis with this difficulty measure was accurately observing the difficulty in SST, we randomly sampled and analysed 100 misclassified data points from SST's test set out of 150 total misclassified. Of these 100, 48 were reviews with both strong positive and negative features and would be difficult for a human to classify, 22 were sarcastic and 8 were mislabelled. The remaining 22 could be easily classified by a human and are misclassified due to errors in the model rather than the data items themselves being difficult to interpret. These findings show that our difficulty measure correctly determined the source of difficulty in SST because 78\% of the errors are implied by our difficulty measure and the remaining 22\% are due to errors in the model itself, not difficulty in the dataset.


\subsection{The Important Areas of Difficulty}
\label{r_areas_of_difficulty}

We hypothesized that the difficulty of a dataset would be determined by four areas not including noise: Class Diversity, Class Balance, Class Interference and Text Complexity. We performed multiple runs of the genetic algorithm, leaving statistics out each time to test which were most important in finding a good difficulty measure which resulted in the following findings:

\paragraph{No Single Characteristic Describes Difficulty} When the Class Diversity statistic was left out of evolution, the highest achieved correlation was $-0.806$, $9\%$ lower than D1 and D2. However, on its own Class Diversity had a correlation of $-0.644$ with model performance. Clearly, Class Diversity is necessary but not sufficient to estimate dataset difficulty. Furthermore, when all measures of Class Diversity and Balance were excluded, the highest achieved correlation was $-0.733$ and when all measures of Class Interference were excluded the best correlation was $-0.727$. These three expected areas of difficulty - Class Diversity, Balance and Interference - must all be measured to get an accurate estimate of difficulty because excluding any of them significantly damages the correlation that can be found. Correlations for each individual statistic are in Table \ref{statistics_corr_table}, in Appendix \ref{app_further_evolution_results}.

\paragraph{Data Complexity Has Little Affect on Difficulty} Excluding all measures of Data Complexity from evolution yielded an average correlation of $-0.869$, only $1\%$ lower than the average when all statistics were included. Furthermore, the only measure of Data Complexity present in D1 and D2 is Distinct Words : Total Words which has a mean value of $0.067$ and therefore contributes very little to the difficulty measure. This shows that while Data Complexity is necessary to achieve top correlation, its significance is minimal in comparison to the other areas of difficulty.

\subsection{Error Analysis}
\label{r_error_analysis}

\subsubsection{Overpowering Class Diversity}

When a dataset has a large number of balanced classes, then Class Diversity dominates the measure. This means that the difficulty measure is not a useful performance estimator for such datasets.

To illustrate this, we created several fake datasets with 1000, 100, 50 and 25 classes. Each dataset had 1000 copies of the same randomly generated string in each class. It was easy for models to overfit and score a 100\% F1 score on these fake datasets.

For the 1000-class fake data, Class Diversity is $6.91$, which by our difficulty measure would indicate that the dataset is extremely difficult. However, all models easily achieve a 100\% F1 score. By testing on these fake datasets, we found that the limit for the number of classes before Class Diversity dominates the difficulty measure and renders it inaccurate is approximately 25. Any datasets with more than 25 classes with an approximately equal number of items per class will be predicted as difficult regardless of whether they actually are because of this diversity measure.

Datasets with more than 25 \textit{unbalanced} classes are still measured accurately. For example, the ATIS dataset \cite{price1990evaluation} has 26 classes but because some of them have only 1 or 2 data items, it is not dominated by Class Diversity. Even when the difficulty measure is dominated by Class Diversity, examining the components of the difficulty measure independently would still be useful as an analysis tool.

\subsubsection{Exclusion of Useful Statistics}

One of our datasets of New Year's Resolution Tweets has 115 classes but only 3507 data points \cite{CrowdFlower}. An ML practitioner knows from the number of classes and data points alone that this is likely to be a difficult dataset for an ML model. 

Our genetic algorithm, based on an unweighted, linear sum, cannot take statistics like data size into account currently because they do not have a convenient range of values; the number of data points in a dataset can vary from several hundred to several million. However, the information is still useful to practitioners in diagnosing the difficulty of a dataset.

Given that the difficulty measure lacks precision and may be better suited to classification than regression as discussed in Section \ref{r_does_generalise}, cannot take account of statistics without a convenient range of values and that the difficulty measure must be interpretable, we suggest that future work could look at combining statistics with a white-box, non-linear algorithm like a decision tree. As opposed to summation, such a combination could take account of statistics with different value ranges and perform either classification or regression while remaining interpretable.

\subsection{How to Reduce the Difficulty Measure}
\label{r_reduce_difficulty}

Here we present some general guidelines on how the four areas of difficulty can be reduced.

Class Diversity can only be sensibly reduced by lowering the number of classes, for example by grouping classes under superclasses. In academic settings where this is not possible, hierarchical learning allows grouping of classes but will produce granular labels at the lowest level \cite{hierDLfortextclass}. Ensuring a large quantity of data in each class will also help models to better learn the features of each class.

Class Interference is influenced by the amount of noise in the data and linguistic phenomena like sarcasm. It can also be affected by the way the data is labelled, for example as shown in Section \ref{r_analysis_tool} where SST has data points with both positive and negative features but only a single label. Filtering noise, restructuring or relabelling ambiguous data points and detecting phenomena like sarcasm will help to reduce class interference. Easily confused classes can also be grouped under one superclass if practitioners are willing to sacrifice granularity to gain performance.

Class Imbalance can be addressed with data augmentation such as thesaurus based methods \cite{charCNN} or word embedding perturbation \cite{augmentingWordEmbs}. Under- and over-sampling can also be utilised \cite{smote} or more data gathered. Another option is transfer learning where knowledge from high data domains can be transferred to those with little data \cite{nluTransferLearning}.

Data Complexity can be managed with large amounts of data. This need not necessarily be labelled - unsupervised pre-training can help models understand the form of complex data before attempting to use it \cite{unreasonableEffectData}. Curriculum learning may also have a similar effect to pre-training \cite{curriculumLearning}.

\subsection{Other Applications of the Measure}

\paragraph{Model Selection} Once the difficulty of a dataset has been calculated, a practitioner can use this to decide whether they need a complex or simple model to learn the data.

\paragraph{Performance Checking and Prediction} Practitioners will be able to compare the results their models get to the scores of other models on datasets of an equivalent difficulty. If their models achieve lower results than what is expected according to the difficulty measure, then this could indicate a problem with the model.

\section{Conclusion} 
\label{conclusion}

When their models do not achieve good results, ML practitioners could potentially calculate thousands of statistics to see what aspects of their datasets are stopping their models from learning. Given this, how do practitioners tell which statistics are the most useful to calculate? Which ones will tell them the most? What changes could they make which will produce the biggest increase in model performance?

In this work, we have presented two measures of text classification dataset difficulty which can be used as analysis tools and performance estimators. We have shown that these measures generalise to unseen datasets. Our recommended measure can be calculated simply by counting the words and labels of a dataset and is formed by adding five different, unweighted statistics together. As the difficulty measure is an unweighted sum, its components can be examined individually to analyse the sources of difficulty in a dataset.

There are two main benefits to this difficulty measure. Firstly, it will reduce the time that practitioners need to spend analysing their data in order to improve model scores. As we have demonstrated which statistics are most indicative of dataset difficulty, practitioners need only calculate these to discover the sources of difficulty in their data. Secondly, the difficulty measure can be used as a performance estimator. When practitioners approach new tasks they need only calculate these simple statistics in order to estimate how well models are likely to perform.

Furthermore, this work has shown that for text classification the areas of Class Diversity, Balance and Interference are essential to measure in order to understand difficulty. Data Complexity is also important, but to a lesser extent.

Future work should firstly experiment with non-linear but interpretable methods of combining statistics into a difficulty measure such as decision trees. Furthermore, it should apply this difficulty measure to other NLP tasks that may require deeper linguistic knowledge than text classification, such as named entity recognition and parsing. Such tasks may require more advanced features than simple word counts as were used in this work.

\newpage

\bibliography{conll2018}
\bibliographystyle{acl_natbib_nourl_conll}

%
%
%
%
%
%
%
%
%
%
%
%
\clearpage

\appendix

\section{Dataset Descriptions}
\label{app_a_dset_desc}
\subsection{Main Datasets}
All 89 datasets used in this paper are available from \url{http://data.wluper.com}.

\paragraph{AG's News Topic Classification Dataset - version 3 (AG)} AG's News Topic Classification Dataset\footnote{\url{https://www.di.unipi.it/~gulli/AG_corpus_of_news_articles.html}} is constructed and used as a benchmark in the paper \cite{charCNN}. It is a collection of more than 1 million news articles gathered from more than 2000 news sources, which is used for research purposes. It has four classes: "Business", "Sci/Tech", "Sports", "World", each class contains 30000 training samples and 1900 testing samples. In total, the training set has 108000 sentences, the validation set has 12000 sentences and the test set has 7600 sentences.

\paragraph{Airline Twitter Sentiment (AT\_Sentiment)} The airline twitter sentiment dataset is crowd-sourced by \cite{CrowdFlower}, it has three sentiment classes to classify the service of the major U.S. airlines: positive, negative, and neutral. The training set has 12444 sentences and the test set has 2196 sentences.

\paragraph{ATIS dataset with Intents (ATIS\_Int)} The airline travel information system dataset \cite{price1990evaluation} is a widely used benchmark dataset for the task of slot filling. The data is from air travel reservation making and follows the (Begin/Inside/Output) format for the semantic label of each word, the class O is assigned for those words without semantic labels. In this paper, we only employed the dataset with intents for the text classification task. In total, the ATIS\_Int data has 26 classes, the training set contains 9956 sentences and the test set contains 893 sentences.

\paragraph{Corporate Messaging (CM)} The Corporate messaging dataset is crowd-sourced by \cite{CrowdFlower}, it has 4 classes about what corporations talk about on social media: "information", "Action", "Dialogue" and "Exclude". The training set has 2650 sentences and the test set has 468 sentences.

\paragraph{Classic Literature Sentence Classification (ClassicLit)} We created this dataset for this work. It consists of The Complete Works of William Shakespeare, War and Peace by Leo Tolstoy, Wuthering Heights by Emily Bront\"{e} and the War of the Worlds by H.G. Wells. Each data item is a sentence from one of these books. All sentences longer than 100 words are discarded. The label of each sentence is which author wrote that sentence. All books were downloaded from the Project Gutenberg\footnote{\url{http://www.gutenberg.org/}} website. There are 40489 training sentences, 5784 validation sentences and 11569 testing sentences.

\paragraph{Classification of Political Social Media (PSM)} The classification of political social media dataset is crowd-sourced by \cite{CrowdFlower}, the social media messages from US Senators and other American politicians are classified into 9 classes ranging from "attack" to "support", the training set has 3500 sentences, the validation set has 500 sentences and the test set has 1000 sentences.  

\paragraph{DBPedia Ontology Classification Dataset - version 2 (DB PEDIA)} The DBpedia dataset are licensed under the terms of GNU Free Documentation License \cite{depdia}, the DBPedia ontology classification dataset is constructed and used as a benchmark in the paper \cite{charCNN}. It has 14 classes, the total size of the training set is 560000 and testing set 70000, we split 10 $\%$ of the training set as validation set with size 5600. Due to the large size of this dataset and the need to increase training speed due to the large number of models we had to train, we randomly sampled 10 $\%$ of the dataset based on the class distribution as our training, validation and test datasets. We later compared our measure to the full, unaltered dataset used by \cite{charCNN} to show generalisation.

\paragraph{Economic News Article Tone and Relevance (ENR)} The Economic News Article Tone and Relevance dataset is crowd-sourced by \cite{CrowdFlower}, it contains classes for whether the article is about the US economy, if so, what tone (1-9) that article is. In this work, we employed a binary classification task by only taking two classes: Yes or No. The training set has 5593 sentences, the validation set has 799 sentences and the test set has 1599 sentences.

\paragraph{Experimental Data for Question Classification (QC)} The Question Classification dataset comes from the \cite{li2002learning}, it classifies questions into six classes: "NUM", "LOC", "HUM", "DESC", "ENTY" and "ABBR", the training set has 4096 sentences, the validation set has 546 sentences and the test set has 500 sentences.

\paragraph{Grammar and Online Product Reviews (GPR)} The Grammar and Online Product Reviews comes from \cite{gprkaggle}, it is a list of reviews of products with 5 classes (rating from 1 to 5). The training set has 49730 sentences, the validation set has 7105 sentences and the test set has 14209 sentences.

\paragraph{Hate Speech (HS)} The hate speech data comes from the work \cite{davidson2017automated} which has three classes: "offensiveLanguage", "hateSpeech", and "neither". The training set has 17348 sentences, the validation set has 2478 sentences and the test set has 1115 sentences.

\paragraph{Large Movie Review Corpus (LMRC)} The large movie review corpus is conducted by \cite{maas2011learning} which contains 50000 reviews from IMDB and an even number of positive and negative reviews. The number of reviews for each movie is not allowed to be more than 30 to avoid correlated ratings. It contains two classes: A negative review has a score $\leq$ 4 out of 10, and a positive review has a score $\geq$ 7 out of 10, no neutral class is included.

\paragraph{London-based Restaurants' Reviews on TripAdvisor (LRR)} The London-based restaurants' reviews on TripAdvisor\footnote{\url{https://www.kaggle.com/PromptCloudHQ/londonbased-restaurants-reviews-on-tripadvisor}} is taken as subset of a bigger dataset (more than 1.8 million restaurants) that was created by extracting data from Tripadvisor.co.uk. It has five classes (rating from 1-5). The training set has 12056 sentences, the validation set has 1722 sentences and the test set has 3445 sentences.

\paragraph{New England Patriots Deflategate sentiment (DFG)} The New England Patriots Deflategate sentiment dataset is crowd-sourced by \cite{CrowdFlower}, it is gathered from Twitter sentiment on chatter around deflated footballs. It has five sentiment classes: negative, slightly negative, neutral, slightly positive and positive. The training set has 8250 sentences, the validation set has 1178 sentences and the test set has 2358 sentences.

\paragraph{New Years Resolutions (NYR)} The 2015 New Year’s resolutions dataset is crowd-sourced by \cite{CrowdFlower}, it contains demographic and geographical data of users and resolution categorizations. In this project, we conducted two text-classification tasks based on different numbers of classes: 10 classes and 115 classes. The training set has 3507 sentences, the validation set has 501 sentences and the test set has 1003 sentences.

\paragraph{Paper Sentence Classification (PSC)} The paper sentence classification dataset comes from the \cite{pscuci}, it contains sentences from the abstract and introduction of 30 articles ranging from biology, machine learning and psychology. There are 5 classes in total, the training set has 2181 sentences, the validation set has 311 sentences and the test set has 625 sentences.

\paragraph{Relevance of terms to disaster relief topics (DT)} The relevance of terms to disaster relief topics dataset is crowd-sourced by \cite{CrowdFlower}, it contains pairs of terms and topics relevant to disaster relief with relevance determinations: relevant and not relevant. The training set has 7597 sentences, the validation set has 1085 sentences and the test set has 2172 sentences.

\paragraph{Review Sentiments (RS)} The review sentiments dataset is generated by 
\cite{kotzias2015group} using an approach from group level labels to instance level labels, 
which is evaluated on three large review datasets: IMDB, Yelp, and Amazon. The dataset contains classes, the training set has 2100 sentences, the validation set has 300 sentences and the test set has 600 sentences.

\paragraph{Self Driving Cars Twitter Sentiment (SDC)} The Self-driving cars Twitter sentiment analysis dataset is crowd-sourced by \cite{CrowdFlower}, it has 6 sentiment classes to classify the sentiments of self driving cars: very positive, slightly positive, neutral, slightly negative, very negative and not relevant. The training set has 6082 sentences and the test set has 1074 sentences.

\paragraph{SMS Spam Collection (SMSS)} The SMS Spam Collection  is a collection of labelled message for mobile phone spam research \cite{smsspam}. It contains two class: spam and ham, the training set has 3901 sentences, the validation set has 558 sentences and the test set has 4957 sentences.  

\paragraph{SNIPS Natural Language Understanding Benchmark (SNIPS)} The SNIPS data is open sourced by \cite{snipsPaper}, it has 7 intents: "AddToPlaylist", "BookRestaurant", "GetWeather", "PlayMusic", "RateBook", "SearchCreativeWork", "SearchScreeningEvent". The training set contains 13784 sentences and the test set contains 700 sentences.

\paragraph{Stanford Sentiment Treebank (SST)} Stanford Sentiment Treebank was introduced by \cite{socher2013recursive}, it is the first corpus with fully labelled parse trees, which is normally used to capture linguistic features and predict the presented compositional semantic effect. It contains 5 sentiment classes: very negative, negative, neutral, positive and very positive. In this project, we conducted two text classification tasks based on a different number of classes in SST: 
\begin{enumerate}
   \item three classes (negative, neutral, positive) classification task, the training data has 236076 sentences, the validation set has 1100 sentences and the test set has 2210 sentences.
   \item binary classification task (negative, positive), the training data has 117220 sentences, the validation set has 872 sentences and the test set has 1821 sentences.
 \end{enumerate}

We used each labelled phrase rather than each sentence as an item of training data which is why the training sets are so large.

\paragraph{Text Emotion Classification (TE)} The Text Emotion Classification dataset is crowd-sourced by \cite{CrowdFlower}, it contains 13 classes for emotional content like happiness or sadness. The training set has 34000 sentences and the test set has 6000 sentences.

\paragraph{Yelp Review Full Star Dataset (YELP)} The Yelp reviews dataset consists of reviews from Yelp, it is extracted from the Yelp Dataset Challenges 2015 data \cite{yelpchallenge}, it is is constructed and used as a benchmark in the paper \cite{charCNN}. In total, there are 650,000 training samples and 50,000 testing samples with 5 classes, we split 10 $\%$ of the training set as validation set. Due to the large size of this dataset and the need to increase training speed due to the large number of models we had to train, we sampled 5 $\%$ of the dataset based on the class distribution as our training, validation and test dataset. We later compare our measure to the full, unaltered dataset as used by \cite{charCNN} to show generalisation.

\paragraph{YouTube Spam Classification (YTS)} The Youtube Spam Classification dataset comes from \cite{ytsuci}, which is a public set of comments collected for spam research. The dataset contains 2 classes, the training set has 1363 sentences, the validation set has 194 sentences and the test set has 391 sentences.

\subsection{Combined Datasets}
\label{a_dset_comb}

We combined the above datasets to make 51 new datasets. The combined datasets were:

\begin{enumerate}
\item AT and ATIS Int
\item AT and ClassicLit
\item AT and CM and DFG and DT and NYR 10 and SDC and TE
\item ATIS Int and ClassicLit
\item ATIS Int and CM
\item ClassicLit and CM
\item ClassicLit and DFG
\item ClassicLit and LMRC
\item ClassicLit and RS
\item ClassicLit and SST
\item ClassicLit and TE
\item CM and AT
\item CM and DFG
\item DFG and AT
\item DFG and ATIS Int
\item DT and AT
\item DT and ATIS Int
\item LMRC and AT
\item LMRC and ATIS Int
\item LMRC and RS and SST
\item LMRC and RS and YTS
\item NYR 10 and AT
\item NYR 10 and ATIS Int
\item NYR 115 and AT
\item NYR 115 and ATIS Int
\item PSC and AT
\item PSC and ATIS Int
\item RS and AT
\item RS and ATIS Int
\item RS and LMRC
\item RS and SST
\item SDC and AT
\item SDC and ATIS Int
\item SNIPS and AT
\item SNIPS and ATIS Int
\item SNIPS and ATIS Int and ClassicLit
\item SNIPS and ATIS Int and PSC
\item SNIPS and ATIS Int and SST
\item SST 2 and AT
\item SST 2 and ATIS Int
\item SST and AT
\item SST and ATIS Int
\item SST and ClassicLit and LMRC
\item SST and LMRC
\item SST and SST 2
\item TE and AT
\item TE and ATIS Int
\item TE and NYR 10
\item YTS and AT
\item YTS and ATIS Int
\item YTS and TE and PSC and RS
\end{enumerate}

\clearpage

\begin{table*}[!t]
\small
\begin{center}
\begin{tabularx}{\linewidth}{|X|X|X|X|}
\hline 
\bf Class Diversity & \bf Class Balance & \bf Class Interference & \bf Data Complexity \\ 
\hline

Shannon Class Diversity    & Class Imbalance & Maximum Unigram Hellinger Similarity       & Distinct Unigrams (Vocab) : Total Unigrams \\
Shannon Class Equitability &                 & Maximum Bigram Hellinger Similarity        & Distinct Bigrams : Total Bigrams           \\
                           &                 & Maximum Trigram Hellinger Similarity       & Distinct Trigrams : Total Trigrams         \\
                           &                 & Maximum 4-gram Hellinger Similarity        & Distinct 4-grams : Total 4-grams           \\
                           &                 & Maximum 5-gram Hellinger Similarity        & Distinct 5-grams : Total 5-grams           \\
                           &                 & Mean Maximum Hellinger Similarity          & Mean Distinct n-grams : Total n-grams      \\
                           &                 & Average Unigram Hellinger Similarity       & Unigram Shannon Diversity                  \\
                           &                 & Average Bigram Hellinger Similarity        & Bigram Shannon Diversity                   \\
                           &                 & Average Trigram Hellinger Similarity       & Trigram Shannon Diversity                  \\   
                           &                 & Average 4-gram Hellinger Similarity        & 4-gram Shannon Diversity                   \\
                           &                 & Average 5-gram Hellinger Similarity        & 5-gram Shannon Diversity                   \\
                           &                 & Mean Average Hellinger Similarity          & Mean n-gram Shannon Diversity              \\
                           &                 & Top Unigram Interference                   & Unigram Shannon Equitability               \\
                           &                 & Top Bigram Interference                    & Bigram Shannon Equitability                \\
                           &                 & Top Trigram Interference                   & Trigram Shannon Equitability               \\
                           &                 & Top 4-gram Interference                    & 4-gram Shannon Equitability                \\
                           &                 & Top 5-gram Interference                    & 5-gram Shannon Equitability                \\
                           &                 & Mean Top n-gram Interference               & Mean n-gram Shannon Equitability           \\
                           &                 & Top Unigram Mutual Information             & Shannon Character Diversity                \\
                           &                 & Top Bigram Mutual Information              & Shannon Character Equitability             \\
                           &                 & Top Trigram Mutual Information             & Inverse Flesch Reading Ease                \\
                           &                 & Top 4-gram Mutual Information              &                                            \\
                           &                 & Top 5-gram Mutual Information              &                                            \\
                           &                 & Mean Top n-gram Mutual Information         &                                            \\

\hline
\end{tabularx}
\end{center}
\caption{\label{statistics_table} The 48 different statistics we calculated of each datasets which we suspected may correlate with model score on different datasets.}
\end{table*}

\FloatBarrier

\section{Dataset Statistic Details}
\label{a_stats}

\subsection{Shannon Diversity Index}
\label{a_stats_sdi}

The Shannon Diversity Index is given by:

\begin{align}
H &= -\sum_{i=1}^{R} p_i ln p_i \\
E_H &= \frac{H}{ln R} \label{eq_shan_equit}
\end{align}

$R$ is the "richness" and corresponds in our case to the number of classes / different n-grams / different characters. $p_i$ is the probability of that class / n-gram / character occurring given by the count base probability distributions described in the method section.

\subsection{Class Imbalance}
\label{a_stats_classimbal}

As part of this paper we presented the class imbalance statistic which is zero if a dataset has precisely the same number of data points in each class. To the best of our knowledge we are the first ones to propose it. Therefore, we will provide a brief mathematical treaty of this metric; in particular we will demonstrate its upper bound. It is defined by:
\begin{align}
Imbal = \sum_{c = 1}^{C} \left| \frac{1}{C} - \frac{n_c}{T_{DATA}} \right|
\end{align}
Where $n_c$ is the count of points in class c, $T_{DATA}$ is the total count of points and $C$ is the total number of classes.
A trivial upper bound can be derived using the triangle inequality and observing that $\sum_{c=1}^{C}\frac{n_c}{T_{DATA}} = 1$ and $0 \leq \frac{n_c}{T_{DATA}} \leq 1$
\begin{align}
(4) &\leq \sum_{c = 1}^{C} \left| \frac{1}{C} \right|+ \left| \frac{n_c}{T_{DATA}} \right| \\
&= C \left| \frac{1}{C} \right|+ \sum_{c=1}^C \left| \frac{n_c}{T_{DATA}} \right| \\
&= C \frac{1}{C} + \sum_{c=1}^C  \frac{n_c}{T_{DATA}} = 1 + 1 = 2 
\end{align}
A tight upper bound, however, is given by: $2(1-\frac{1}{C})$, i.e. when one class has all the data points, while all the other classes have zero data points. A brief derivation hereof goes as follows: \\
We let $\frac{n_c}{T_{DATA}}=p_c\in [0,1]$, with $\sum_c p_c = 1$ then
\begin{align}
(4) = \sum_{c = 1}^{C} \left| \frac{1}{C} - p_c \right|
\end{align}
We will find the upper bound by maximising (7). Now we split the p's into two sets. Set $I$, consists of all the $p_i\geq \frac{1}{C}$ and set $J$, which consists of $p_j < \frac{1}{C}$. Then:
\begin{align}
(7) &= \sum_{i \in I} \left| \frac{1}{C} - p_i \right| + \sum_{j \in J} \left| \frac{1}{C} - p_j \right| \\
&= \sum_{i \in I} (p_i - \frac{1}{C}) + \sum_{j \in J} (\frac{1}{C} - p_j) \\
&= \sum_i p_i + \sum_j \frac{1}{C} - \sum_j p_j - \sum_i \frac{1}{C}
\end{align}
As we are free to choose the p's, it is clear from (10) that all the p's in set $J$ should be 0, as they are a negative term in the sum. This leaves us with:
\begin{align}
(11) \leq \sum_i p_i + \sum_j \frac{1}{C} - \sum_i \frac{1}{C}
\end{align}
Where $\sum_i p_i = 1$, as $\sum_i p_i + \sum_j p_j =1$, hence:
\begin{align}
(12) = 1 + \sum_j \frac{1}{C} - \sum_i \frac{1}{C}
\end{align}
Now the only term which can be maximised is the split between the set $I$ and $J$, leaving us with the maximum the Imbalance can achieve, i.e. split of $|I|=1$ and $|J|=C-1$ and therefore:
\begin{align}
(13) \leq 1 + (C-1) \frac{1}{C} - \frac{1}{C} = 2 (1-\frac{1}{C}) \qed
\end{align}
This shows that the Imbalance metric aligns with our intuitive definition of imbalance. 

\subsection{Hellinger Similarity}
\label{a_stats_helldist}

The Hellinger Similarity is given by:

\begin{align}
HS(P, Q) = 1 - \frac{1}{\sqrt{2}} \sqrt{\sum_{i=1}^{k} (\sqrt{p_i} - \sqrt{q_i})^2}
\end{align}

Where $p_i$ is the probability of word $i$ occurring and $k$ is the number of words in either $P$ or $Q$, whichever is larger.

\subsection{Jaccard Distance}
\label{a_stats_jaccarddist}

The Jaccard distance is given by:

\begin{align}
J(P, Q) = \frac{\left| P \cap Q \right|}{\left| P \cup Q \right|}
\end{align}

Where P is the set of words in one class and Q is the set of words in another.

\subsection{Mutual Information}
\label{a_stats_minfo}

The mutual information statistic between two classes $X$ and $Y$ is:

\begin{align}
MI = \sum_{x \in X} \sum_{y \in Y} p(x, y) log \left( \frac{p(x, y)}{p(x) p(y)} \right) 
\end{align}

$p(x)$ represents the probability of a word occurring in class $X$. $p(x, y)$ is the sum of the count of the word in class $X$ and the count of the word in class $Y$ divided by the sum of the total count of words in both classes.

\clearpage

\section{Model Descriptions}
\label{app_b_model_desc}

All of our models were implemented in Python with TensorFlow\footnote{\url{https://www.tensorflow.org/}} for neural networks and scikit-learn\footnote{\url{http://scikit-learn.org/}} for classical models.

\subsection{Word Embedding Based Models}
Five of our models use word embeddings. We use a FastText \cite{fasttextRef} model trained on the One Billion Word Corpus \cite{billionWordCorpus} which gives us an open vocabulary. Embedding size is 128. All models use a learning rate of 0.001 and the Adam optimiser \cite{adamOptRef}. The loss function is weighted cross entropy where each class is weighted according to the proportion of its occurrence in the data. This allows the models to handle unbalanced data.

\paragraph{LSTM- and GRU-based RNNs} Two simple recurrent neural networks (RNNs) with LSTM cells \cite{lstmRef} and GRU cells \cite{gruRef}, which are directly passed the word embeddings. The cells have 64 hidden units and no regularisation. The final output is projected and passed through a softmax to give class probabilities.

\paragraph{Bidirectional LSTM and GRU RNNs} Uses the bidirectional architecture proposed by \cite{bilstmRef}, the networks are directly passed the word embeddings. The cells have 64 units each. The final forward and backward states are concatenated giving a 128 dimensional vector which is projected and passed through a softmax to give class probabilities.

\paragraph{Multi-Layer Perceptron (MLP)} A simple single hidden layer neural network with 64 hidden units and ReLU non-linearity. The word embeddings for each word of the sentence are summed before being passed through the network, following the approach of \cite{vectorSumRef}.

\subsection{TF-IDF Based Models}
Six of our models use TFIDF based vectors \cite{tfidfRef} as input. Loss is weighted to handle class imbalance, but the function used varies between models.

\paragraph{Adaboost} An ensemble algorithm which trains many weak learning models on subsets of the data using the SAMME algorithm \cite{adaboostsammeRef}. We use a learning rate of 1 and 100 different weak learners. The weak learners are decision trees which measure split quality with the Gini impurity and have a maximum depth of 5.

\paragraph{Gaussian Naive Bayes} A simple and fast algorithm based on Bayes' theorem with the naive independence assumption among features. It can be competitive with more advanced models like Support Vector Machines with proper data preprocessing \cite{naiveBayesvsSVMref}. We do not preprocess the data in any way and simply pass the TF-IDF vector.

\paragraph{k-Nearest Neighbors} A non-parametric model which classifies new data by the majority class among the k nearest neighbors of each new item. We set $k = 5$. Euclidean distance is used as the the distance metric between items.

\paragraph{Logistic Regression} A regression model which assigns probabilities to each class. For multiple classes this becomes multinomial logistic or softmax regression.  We use L2 regularization with this model with regularization strength set to 1.

\paragraph{Random Forest} An ensemble method that fits many decision trees on subsamples of the dataset. The forest then chooses the best classification over all trees. Each of our forests contain 100 trees and have a maximum depth of 100. Trees require a minimum of 25 samples to create a leaf node.

\paragraph{Support Vector Machine (SVM)} A mathematical method which maximises the distance between the decision boundary and the support vectors of each class. Implemented with a linear kernel, l2 regularisation with a strength of 1 and squared hinge loss. Multi-class is handled by training multiple binary classifiers in a "one vs rest" scheme as followed by \cite{svmonevrest}. Due to the inefficiency of SVMs operating on large datasets, all datasets are sub sampled randomly to $10000$ data items before fitting this model.

\subsection{Character Based Models}
We use a single character based model which is a convolutional neural network (CNN) inspired by \cite{charCNN}. Similar to our word embedding based models, it uses a learning rate of 0.001 and Adam optimizer. It receives one-hot representations of the characters and embeds these into a 128-dimensional dense representation. It then has three convolutional layers of kernel sizes $5$, $3$ and $3$ with channel sizes of $256$, $128$ and $64$ respectively. All stride sizes are 1 and valid padding is used. The convolutional layers are followed by dropout with probability set to 0.5 during training and 1 during testing. The dropout is followed by two hidden layers with ReLU non-linearity with $1000$ then $256$ hidden units respectively. Finally this output is projected and passed through a softmax layer to give class probabilities.

\clearpage

\section{Further Evolution Results}
\label{app_further_evolution_results}

\subsection{Other Discovered Difficulty Measures}

\begin{figure*}[!htb]
\includegraphics[width=\linewidth]{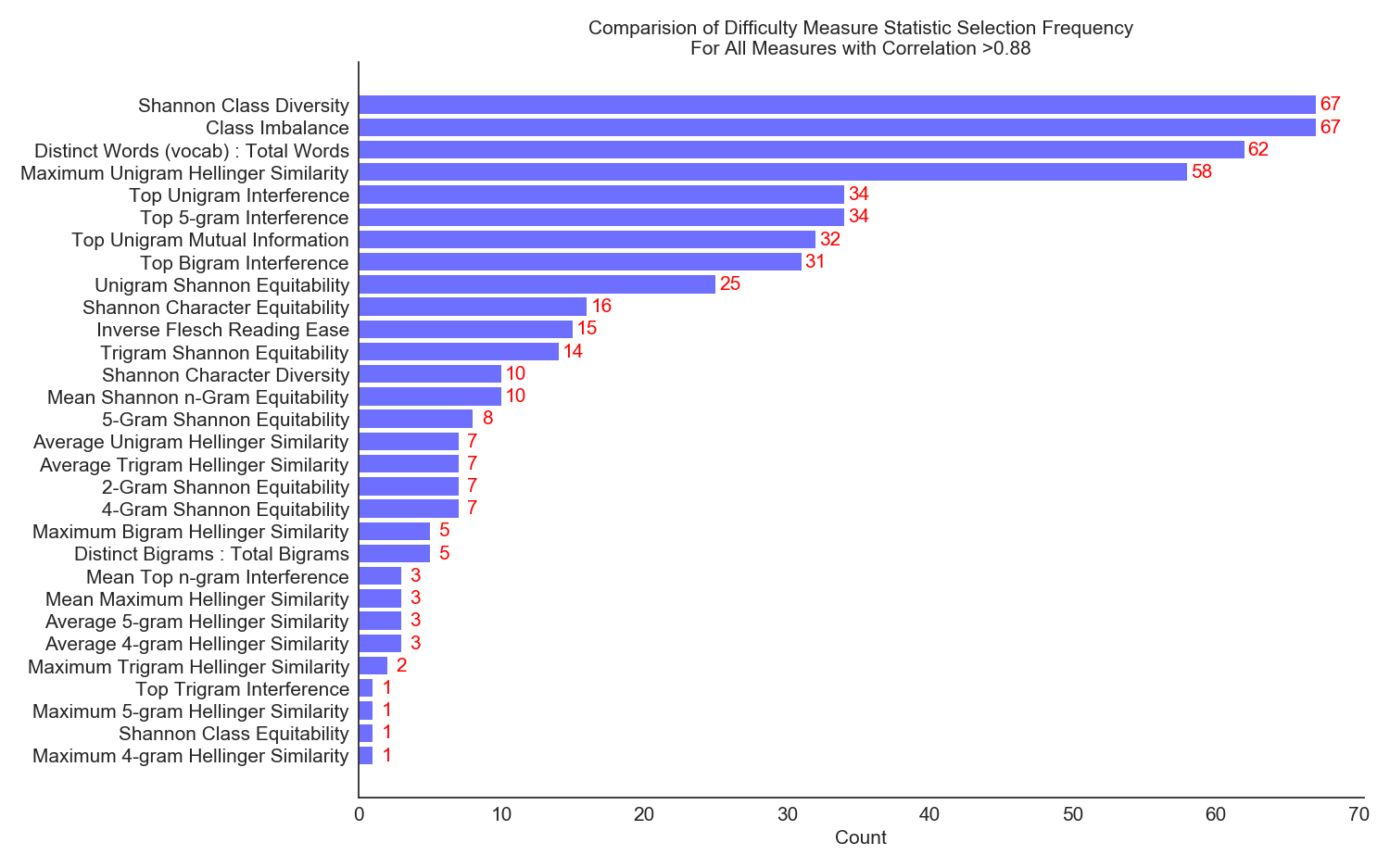}
\caption{\label{fig_selection_freq} Frequency of different statistics' selection during evolution for all difficulty measures found with correlation greater than 0.88.}
\end{figure*}

This appendix contains a listing of other difficulty measures discovered by the evolutionary process. The selection frequencies of different components are visualised in Figure \ref{fig_selection_freq}.

\begin{itemize}
\item \textit{Distinct Words (vocab) : Total Words + Top 5-gram Interference + Class Imbalance + Shannon Class Diversity + Maximum Unigram Hellinger Similarity + Top Unigram Mutual Information} : 0.884524539285
\item \textit{Distinct Words (vocab) : Total Words + Top 5-gram Interference + Class Imbalance + Shannon Class Diversity + Unigram Shannon Equitability + Maximum Unigram Hellinger Similarity + Maximum 4-gram Hellinger Similarity + Top Unigram Mutual Information + Shannon Character Equitability} : 0.884444632266
\item \textit{Distinct Words (vocab) : Total Words + Top 5-gram Interference + Class Imbalance + Shannon Class Diversity + Average 4-gram Hellinger Similarity + Maximum Unigram Hellinger Similarity + Top Unigram Mutual Information + Shannon Character Equitability} : 0.884417837328
\item \textit{Distinct Words (vocab) : Total Words + Top 5-gram Interference + Class Imbalance + Shannon Class Diversity + Unigram Shannon Equitability + Trigram Shannon Equitability + Maximum Unigram Hellinger Similarity + Top Unigram Mutual Information} : 0.884389247487
\item \textit{Distinct Words (vocab) : Total Words + Top 5-gram Interference + Class Imbalance + Shannon Class Diversity + Unigram Shannon Equitability + Trigram Shannon Equitability + Maximum Unigram Hellinger Similarity + Top Unigram Mutual Information + Inverse Flesch Reading Ease} : 0.883771907471
\item \textit{Distinct Words (vocab) : Total Words + Top Unigram Interference + Top Bigram Interference + Top 5-gram Interference + Class Imbalance + Shannon Class Diversity + Unigram Shannon Equitability + 4-Gram Shannon Equitability + Maximum Unigram Hellinger Similarity + Shannon Character Equitability + Inverse Flesch Reading Ease} : 0.883726094418
\item \textit{Distinct Words (vocab) : Total Words + Top 5-gram Interference + Class Imbalance + Shannon Class Diversity + Average 5-gram Hellinger Similarity + Maximum Unigram Hellinger Similarity + Top Unigram Mutual Information} : 0.883722932273
\item \textit{Distinct Words (vocab) : Total Words + Top 5-gram Interference + Class Imbalance + Shannon Class Diversity + Mean Shannon n-Gram Equitability + Maximum Unigram Hellinger Similarity + Top Unigram Mutual Information + Shannon Character Diversity} : 0.883590459193
\item \textit{Distinct Words (vocab) : Total Words + Top Unigram Interference + Top Bigram Interference + Top 5-gram Interference + Class Imbalance + Shannon Class Diversity + Unigram Shannon Equitability + Trigram Shannon Equitability + 5-Gram Shannon Equitability + Maximum Unigram Hellinger Similarity} : 0.883561148159
\item \textit{Distinct Words (vocab) : Total Words + Top 5-gram Interference + Class Imbalance + Shannon Class Diversity + 2-Gram Shannon Equitability + 5-Gram Shannon Equitability + Maximum Unigram Hellinger Similarity + Top Unigram Mutual Information} : 0.883335585611
\item \textit{Distinct Words (vocab) : Total Words + Top Unigram Interference + Top Bigram Interference + Top 5-gram Interference + Class Imbalance + Shannon Class Diversity + 2-Gram Shannon Equitability + 4-Gram Shannon Equitability + Maximum Unigram Hellinger Similarity + Shannon Character Equitability + Inverse Flesch Reading Ease} : 0.883257036313
\item \textit{Distinct Words (vocab) : Total Words + Top 5-gram Interference + Class Imbalance + Shannon Class Diversity + Average Trigram Hellinger Similarity + Maximum Unigram Hellinger Similarity + Top Unigram Mutual Information + Shannon Character Equitability + Inverse Flesch Reading Ease} : 0.883217937163
\item \textit{Distinct Words (vocab) : Total Words + Top 5-gram Interference + Class Imbalance + Shannon Class Diversity + Maximum Unigram Hellinger Similarity + Top Unigram Mutual Information + Shannon Character Diversity + Inverse Flesch Reading Ease} : 0.883092632656
\item \textit{Distinct Words (vocab) : Total Words + Class Imbalance + Shannon Class Diversity + Unigram Shannon Equitability + Maximum Unigram Hellinger Similarity + Maximum Trigram Hellinger Similarity + Top Unigram Mutual Information} : 0.882946516641
\item \textit{Distinct Words (vocab) : Total Words + Top Unigram Interference + Top Bigram Interference + Top 5-gram Interference + Class Imbalance + Shannon Class Diversity + Unigram Shannon Equitability + Trigram Shannon Equitability + Mean Shannon n-Gram Equitability + Maximum Unigram Hellinger Similarity + Shannon Character Equitability + Inverse Flesch Reading Ease} : 0.882914430188
\item \textit{Distinct Words (vocab) : Total Words + Top Unigram Interference + Top Bigram Interference + Top 5-gram Interference + Class Imbalance + Shannon Class Diversity + Mean Shannon n-Gram Equitability + Maximum Unigram Hellinger Similarity} : 0.882863026072
\item \textit{Distinct Bigrams : Total Bigrams + Top Unigram Interference + Top Bigram Interference + Top 5-gram Interference + Class Imbalance + Shannon Class Diversity + Unigram Shannon Equitability + Maximum Unigram Hellinger Similarity} : 0.882825047536
\item \textit{Distinct Words (vocab) : Total Words + Top Unigram Interference + Top Bigram Interference + Top 5-gram Interference + Class Imbalance + Shannon Class Diversity + Trigram Shannon Equitability + Maximum Unigram Hellinger Similarity} : 0.882628270942
\item \textit{Distinct Bigrams : Total Bigrams + Top Unigram Interference + Top Bigram Interference + Top 5-gram Interference + Class Imbalance + Shannon Class Diversity + Unigram Shannon Equitability + Maximum Unigram Hellinger Similarity + Shannon Character Equitability} : 0.882562918832
\item \textit{Distinct Words (vocab) : Total Words + Class Imbalance + Shannon Class Diversity + Mean Shannon n-Gram Equitability + Average Trigram Hellinger Similarity + Maximum Unigram Hellinger Similarity + Top Unigram Mutual Information + Shannon Character Equitability} : 0.882376082243
\item \textit{Distinct Words (vocab) : Total Words + Top Unigram Interference + Top Bigram Interference + Class Imbalance + Shannon Class Diversity + Unigram Shannon Equitability + Maximum Unigram Hellinger Similarity + Shannon Character Equitability} : 0.882297072242
\item \textit{Distinct Words (vocab) : Total Words + Top 5-gram Interference + Class Imbalance + Shannon Class Diversity + 2-Gram Shannon Equitability + Average Trigram Hellinger Similarity + Maximum Unigram Hellinger Similarity + Top Unigram Mutual Information + Inverse Flesch Reading Ease} : 0.882245884638
\item \textit{Distinct Words (vocab) : Total Words + Top Unigram Interference + Top Bigram Interference + Top 5-gram Interference + Class Imbalance + Shannon Class Diversity + Unigram Shannon Equitability + 2-Gram Shannon Equitability + Average Unigram Hellinger Similarity} : 0.882237171884
\item \textit{Distinct Words (vocab) : Total Words + Class Imbalance + Shannon Class Diversity + Average 5-gram Hellinger Similarity + Maximum Unigram Hellinger Similarity + Top Unigram Mutual Information + Shannon Character Equitability} : 0.882046043522
\item \textit{Distinct Words (vocab) : Total Words + Top 5-gram Interference + Class Imbalance + Shannon Class Diversity + Unigram Shannon Equitability + Trigram Shannon Equitability + Maximum Unigram Hellinger Similarity + Top Unigram Mutual Information + Shannon Character Diversity + Inverse Flesch Reading Ease} : 0.881936634248
\item \textit{Distinct Words (vocab) : Total Words + Top Unigram Interference + Top Bigram Interference + Class Imbalance + Shannon Class Diversity + Unigram Shannon Equitability + Average Unigram Hellinger Similarity} : 0.881760155361
\item \textit{Distinct Words (vocab) : Total Words + Top 5-gram Interference + Class Imbalance + Shannon Class Diversity + Unigram Shannon Equitability + 5-Gram Shannon Equitability + Maximum Bigram Hellinger Similarity + Maximum Trigram Hellinger Similarity + Top Unigram Mutual Information} : 0.881643119225
\item \textit{Distinct Words (vocab) : Total Words + Top 5-gram Interference + Class Imbalance + Shannon Class Diversity + Maximum Unigram Hellinger Similarity + Maximum Bigram Hellinger Similarity + Top Unigram Mutual Information} : 0.881581392494
\item \textit{Distinct Words (vocab) : Total Words + Top 5-gram Interference + Class Imbalance + Shannon Class Diversity + Trigram Shannon Equitability + Maximum Unigram Hellinger Similarity + Maximum 5-gram Hellinger Similarity + Top Unigram Mutual Information} : 0.881517499763
\item \textit{Distinct Words (vocab) : Total Words + Top Unigram Interference + Top Bigram Interference + Class Imbalance + Shannon Class Diversity + Unigram Shannon Equitability + 4-Gram Shannon Equitability + Mean Shannon n-Gram Equitability + Maximum Unigram Hellinger Similarity} : 0.881490427511
\item \textit{Distinct Words (vocab) : Total Words + Class Imbalance + Shannon Class Diversity + Trigram Shannon Equitability + Average Trigram Hellinger Similarity + Maximum Unigram Hellinger Similarity + Top Unigram Mutual Information} : 0.88147122781
\item \textit{Distinct Words (vocab) : Total Words + Top 5-gram Interference + Class Imbalance + Shannon Class Diversity + 4-Gram Shannon Equitability + Average 5-gram Hellinger Similarity + Maximum Unigram Hellinger Similarity + Top Unigram Mutual Information + Shannon Character Diversity + Inverse Flesch Reading Ease} : 0.881440344906
\item \textit{Distinct Bigrams : Total Bigrams + Top Unigram Interference + Top Bigram Interference + Top 5-gram Interference + Class Imbalance + Shannon Class Diversity + Maximum Unigram Hellinger Similarity} : 0.881440337829
\item \textit{Distinct Words (vocab) : Total Words + Top Unigram Interference + Top Bigram Interference + Class Imbalance + Shannon Class Diversity + Mean Shannon n-Gram Equitability + Maximum Unigram Hellinger Similarity} : 0.881426394865
\item \textit{Distinct Words (vocab) : Total Words + Top Unigram Interference + Top Bigram Interference + Class Imbalance + Shannon Class Diversity + Trigram Shannon Equitability + 4-Gram Shannon Equitability + Maximum Unigram Hellinger Similarity + Shannon Character Equitability} : 0.881404209076
\item \textit{Distinct Words (vocab) : Total Words + Class Imbalance + Shannon Class Diversity + Maximum Unigram Hellinger Similarity + Top Unigram Mutual Information} : 0.881365728443
\item \textit{Distinct Words (vocab) : Total Words + Top Unigram Interference + Top Bigram Interference + Class Imbalance + Shannon Class Diversity + 5-Gram Shannon Equitability + Mean Shannon n-Gram Equitability + Maximum Unigram Hellinger Similarity} : 0.881342679515
\item \textit{Distinct Words (vocab) : Total Words + Class Imbalance + Shannon Class Diversity + Maximum Unigram Hellinger Similarity + Mean Maximum Hellinger Similarity + Top Unigram Mutual Information} : 0.881340929845
\item \textit{Distinct Words (vocab) : Total Words + Class Imbalance + Shannon Class Diversity + Unigram Shannon Equitability + Maximum Unigram Hellinger Similarity + Top Unigram Mutual Information + Shannon Character Diversity} : 0.88117529932
\item \textit{Distinct Words (vocab) : Total Words + Top Unigram Interference + Top Bigram Interference + Class Imbalance + Shannon Class Diversity + Trigram Shannon Equitability + Maximum Unigram Hellinger Similarity + Shannon Character Diversity} : 0.881163020765
\item \textit{Distinct Words (vocab) : Total Words + Top Unigram Interference + Top Trigram Interference + Top 5-gram Interference + Class Imbalance + Shannon Class Diversity + Unigram Shannon Equitability + Maximum Unigram Hellinger Similarity} : 0.881044616332
\item \textit{Distinct Words (vocab) : Total Words + Top Unigram Interference + Top Bigram Interference + Class Imbalance + Shannon Class Diversity + 4-Gram Shannon Equitability + Mean Shannon n-Gram Equitability + Average Unigram Hellinger Similarity} : 0.88091437587
\item \textit{Distinct Words (vocab) : Total Words + Top Unigram Interference + Top Bigram Interference + Mean Top n-gram Interference + Class Imbalance + Shannon Class Diversity + Unigram Shannon Equitability + Maximum Unigram Hellinger Similarity + Shannon Character Diversity + Shannon Character Equitability} : 0.880910807679
\item \textit{Distinct Words (vocab) : Total Words + Top Unigram Interference + Top Bigram Interference + Class Imbalance + Shannon Class Diversity + Trigram Shannon Equitability + Average Unigram Hellinger Similarity} : 0.880885142235
\item \textit{Distinct Words (vocab) : Total Words + Top Unigram Interference + Top Bigram Interference + Top 5-gram Interference + Class Imbalance + Shannon Class Diversity + 5-Gram Shannon Equitability + Average Unigram Hellinger Similarity} : 0.880837764104
\item \textit{Distinct Words (vocab) : Total Words + Class Imbalance + Shannon Class Diversity + Maximum Unigram Hellinger Similarity + Maximum Bigram Hellinger Similarity + Top Unigram Mutual Information} : 0.880709196549
\item \textit{Distinct Words (vocab) : Total Words + Top Unigram Interference + Top 5-gram Interference + Mean Top n-gram Interference + Class Imbalance + Shannon Class Diversity + Unigram Shannon Equitability + Maximum Unigram Hellinger Similarity + Shannon Character Diversity} : 0.880654042756
\item \textit{Distinct Words (vocab) : Total Words + Top Unigram Interference + Top 5-gram Interference + Mean Top n-gram Interference + Class Imbalance + Shannon Class Diversity + Unigram Shannon Equitability + 5-Gram Shannon Equitability + Mean Shannon n-Gram Equitability + Maximum Unigram Hellinger Similarity + Inverse Flesch Reading Ease} : 0.88058845366
\item \textit{Distinct Bigrams : Total Bigrams + Top Unigram Interference + Top Bigram Interference + Class Imbalance + Shannon Class Diversity + Unigram Shannon Equitability + Maximum Unigram Hellinger Similarity + Maximum Bigram Hellinger Similarity} : 0.88057312013
\item \textit{Distinct Words (vocab) : Total Words + Top Unigram Interference + Top Bigram Interference + Class Imbalance + Shannon Class Diversity + 2-Gram Shannon Equitability + 5-Gram Shannon Equitability + Maximum Unigram Hellinger Similarity + Shannon Character Diversity + Shannon Character Equitability} : 0.880527649949
\item \textit{Distinct Words (vocab) : Total Words + Top Unigram Interference + Top Bigram Interference + Class Imbalance + Shannon Class Diversity + 2-Gram Shannon Equitability + Trigram Shannon Equitability + Maximum Unigram Hellinger Similarity + Inverse Flesch Reading Ease} : 0.880466229085
\item \textit{Distinct Words (vocab) : Total Words + Top Unigram Interference + Top Bigram Interference + Class Imbalance + Shannon Class Diversity + Unigram Shannon Equitability + Average Unigram Hellinger Similarity + Inverse Flesch Reading Ease} : 0.880427741747
\item \textit{Distinct Words (vocab) : Total Words + Top Unigram Interference + Top Bigram Interference + Class Imbalance + Shannon Class Diversity + 5-Gram Shannon Equitability + Maximum Unigram Hellinger Similarity} : 0.880408773828
\item \textit{Distinct Words (vocab) : Total Words + Top Unigram Interference + Top Bigram Interference + Top 5-gram Interference + Class Imbalance + Shannon Class Diversity + Unigram Shannon Equitability + Average 4-gram Hellinger Similarity + Maximum Unigram Hellinger Similarity} : 0.880334639215
\item \textit{Distinct Words (vocab) : Total Words + Class Imbalance + Shannon Class Diversity + Mean Shannon n-Gram Equitability + Average Trigram Hellinger Similarity + Maximum Unigram Hellinger Similarity + Top Unigram Mutual Information + Inverse Flesch Reading Ease} : 0.880326776791
\item \textit{Distinct Words (vocab) : Total Words + Class Imbalance + Shannon Class Diversity + Unigram Shannon Equitability + Trigram Shannon Equitability + Maximum Unigram Hellinger Similarity + Mean Maximum Hellinger Similarity + Top Unigram Mutual Information} : 0.880295177778
\item \textit{Distinct Words (vocab) : Total Words + Top Unigram Interference + Top Bigram Interference + Class Imbalance + Shannon Class Diversity + Maximum Unigram Hellinger Similarity} : 0.880252374975
\item \textit{Distinct Words (vocab) : Total Words + Class Imbalance + Shannon Class Diversity + Trigram Shannon Equitability + Average Trigram Hellinger Similarity + Maximum Unigram Hellinger Similarity + Top Unigram Mutual Information + Shannon Character Diversity} : 0.880239646699
\item \textit{Distinct Words (vocab) : Total Words + Top Unigram Interference + Top Bigram Interference + Class Imbalance + Shannon Class Diversity + 2-Gram Shannon Equitability + Average Unigram Hellinger Similarity + Inverse Flesch Reading Ease} : 0.880200393627
\item \textit{Distinct Words (vocab) : Total Words + Class Imbalance + Shannon Class Diversity + 4-Gram Shannon Equitability + Average Trigram Hellinger Similarity + Maximum Unigram Hellinger Similarity + Top Unigram Mutual Information + Shannon Character Equitability + Inverse Flesch Reading Ease} : 0.880083849581
\end{itemize}

\clearpage

\begin{table*}[!t]
\small
\begin{center}
\begin{tabularx}{\linewidth}{|X|X|}
\hline 
\bf Statistic & \bf Correlation \\ 
\hline

Maximum Unigram Hellinger Similarity      & 0.720896895887 \\
Top Unigram Interference                                 & 0.64706340007 \\
Maximum Bigram Hellinger Similarity      & 0.619410655023 \\
Mean Maximum Hellinger Similarity & 0.599742584859 \\
Mean Top N-Gram Interference                             & 0.592624636419 \\
Average Unigram Hellinger Similarity      & 0.574120851308 \\
Top Bigram Interference                                 & 0.574018328147 \\
Top Trigram Interference                                 & 0.556869160804 \\
Shannon Class Diversity                                 & 0.495247387609 \\
Maximum Trigram Hellinger Similarity     & 0.470443549996 \\
Top 5-Gram Interference                                 & 0.469209823975 \\
Average Bigram Hellinger Similarity      & 0.457163222902 \\
Mean Average Hellinger Similarity & 0.454790305987 \\
Top 4-Gram Interference                                 & 0.418374832964 \\
Maximum Unigram Hellinger Similarity     & 0.332573671726 \\
Average Trigram Hellinger Similarity      & 0.328687842958 \\
Top Unigram Mutual Information             & 0.293673742958 \\
Maximum 5-gram Hellinger Similarity      & 0.261369081098 \\
Average 4-gram Hellinger Similarity      & 0.24319918737 \\
Average 5-gram Hellinger Similarity     & 0.20741866152 \\
Top 5-gram Mutual Information             & 0.18246852683 \\
Class Imbalance                                         & 0.164274169881 \\
Mean n-Gram Shannon Equitability                         & 0.14924393263 \\
4-Gram Shannon Equitability                             & 0.142930086195 \\
Trigram Shannon Equitability                             & 0.130883685416 \\
Unigram Shannon Equitability                             & 0.129571167512 \\
5-Gram Shannon Equitability                             & 0.118068879785 \\
Bigram Shannon Equitability                             & 0.116996612078 \\
Unigram Shannon Diversity                                & 0.0587541973146 \\
Distinct Words (Vocab) : Total Words                    & 0.0578981403589 \\
Bigram Shannon Diversity                                & 0.0516963177593 \\
Mean n-Gram Shannon Diversity                            & 0.0440696293705 \\
Shannon Character Diversity                                  & 0.0431234569786 \\
Mean Top n-gram Mutual Information                     & 0.0413350594379 \\
Shannon Character Equitability                               & 0.0402159715373 \\
Trigram Shannon Diversity                                & 0.0396008851652 \\
Shannon Class Equitability                              & 0.0360726401633 \\
Top Trigram Mutual Information             & 0.0337710575222 \\
Top 4-gram Mutual Information             & 0.0279796567333 \\
4-Gram Shannon Diversity                                & 0.0259739834385 \\
Top Bigram Mutual Information             & 0.0257123532616 \\
Distinct Bigrams : Total Bigrams                                          & 0.0252155036575 \\
Inverse Flesch Reasing Ease                             & 0.0250329647438 \\
5-Gram Shannon Diversity                                & 0.0189276868112 \\
Mean Distinct n-grams : Total n-grams                                      & 0.0141636605848 \\
Distinct 5-grams : Total 5-grams                                          & 0.00664063690957 \\
Distinct Trigrams : Total Trigrams                                          & 0.00465734012651 \\
Distinct 4-grams : Total 4-grams                                          & 0.0015168555015 \\

 \hline
\end{tabularx}
\end{center}
\caption{\label{statistics_corr_table} The 48 different statistics we calculated and their correlations with model score across datasets}
\end{table*}
 
\end{document}